\documentclass{article}

\usepackage[utf8]{inputenc} 
\usepackage[T1]{fontenc} 
\usepackage{hyperref} 
\usepackage{url} 
\usepackage{booktabs} 
\usepackage{amsfonts} 
\usepackage{nicefrac} 
\usepackage{microtype} 

\usepackage{booktabs} 

\usepackage[ruled]{algorithm2e} 

\usepackage{algorithmic}
\usepackage{algorithm2e}
\SetAlFnt{\small}
\SetAlCapFnt{\small}
\SetAlCapNameFnt{\small}
\SetAlCapHSkip{0pt}
\IncMargin{-\parindent}

\usepackage{helvet} 
\usepackage{courier} 
\usepackage{url} 
\usepackage{graphicx} 
\usepackage{booktabs}
\usepackage{amsmath}

\usepackage{multirow}
\usepackage{multirow}
\usepackage{makecell}
\usepackage{caption}
\usepackage{subfig}
\usepackage{soul}
\usepackage{color}

\usepackage[preprint,nonatbib]{nips_2019}

\title{Node Attribute Generation on Graphs}

\author{%
 Xu Chen \quad
  Siheng Chen \quad
  Huangjie Zheng \quad
  Jiangchao Yao 
 \And
  Kenan Cui \quad
  Ya Zhang \quad
  Ivor W. Tsang
  \AND
 \texttt{{\tt \{xuchen2016, conancui, ya\_zhang\}@sjtu.edu.cn}}\\
 \texttt{sihengc@andrew.cmu.edu} \quad
 \texttt{huangjie.zheng@utexas.edu} \\
 \texttt{jiangchao.yjc@alibaba-inc.com} \quad
 \texttt{ivor.tsang@uts.edu.au} \\
}

\begin{document}
\maketitle
\begin{abstract}
Graph structured data provide two-fold information: graph structures and node attributes. Numerous graph-based algorithms rely on both information to achieve success in supervised tasks, such as node classification and link prediction. However, node attributes could be missing or incomplete, which significantly deteriorates the performance. The task of node attribute generation aims to generate attributes for those nodes whose attributes are completely unobserved. This task benefits many real-world problems like profiling, node classification and graph data augmentation. To tackle this task, we propose a deep adversarial learning based method to generate node attributes; called node attribute neural generator (NANG). NANG learns a unifying latent representation which is shared by both node attributes and graph structures and can be translated to different modalities. We thus use this latent representation as a bridge to convert information from one modality to another. We further introduce practical applications to quantify the performance of node attribute generation. Extensive experiments are conducted on four real-world datasets and the empirical results show that node attributes generated by the proposed method are high-qualitative and beneficial to other applications. The datasets and codes are available online\footnote{https://drive.google.com/file/d/15C8vToeVFXrNvkqL2-Zn43j3wq1edTfq/view?usp=sharing}.
\end{abstract}

\section{Introduction}
In recent years, representation learning on graphs has attracted great research interest because of its great potential in numerous machine learning tasks, such as semi-supervised learning~\cite{kipf2016semi} and zero/few-shot learning~\cite{fu2015transductive,garcia2017few}, as well as numerous real-world applications, such as recommendation systems~\cite{ying2018graph} and action recognition~\cite{li2019actional}. A standard supervised-learning paradigm on graph is to predict some class information based on observed graph structures and node attributes. For example,~\cite{kipf2016semi,kipf2016variational} proposed graph convolutional neural networks to learn graph embeddings based on both graph structures and node attributes. In practice, either graph structures or node attributes could be inaccurate, incomplete or missing, which would cause significant deterioration to the performance. To make more complete description about graph structures,~\cite{you2018graphrnn,bojchevski2018netgan} consider graph structure generation by using random walks and Recurrent Neural Networks (RNNs).

In this paper, we focus on generating node attributes for missing data. Specifically, in a fixed and given graph structure, only partial nodes have attributes and attributes of the rest nodes are completely missing and inaccessible. We aim to generate attributes for those nodes that do not have any observed attribute. Compared to the completion problem~\cite{kalofolias2014matrix} where all nodes could have partially observed attributes, in our task, attributes of some nodes are completely unknown, we thus call it~\emph{ node attribute generation}. This task is related to numerous real-world challenges. For example, in citation networks, raw attributes and detailed descriptions of papers may be missing due to the copyright protection. In item co-occurrence graph, descriptive tags for items may be missing because of the expensive tagging labour. The task of node attribute generation not only retrieves unknown information, but also benefits many supervised-learning tasks, such as profiling and node classification. The generated node attributes can also serve as additional inputs to improve subsequent tasks.

Recently, a series of deep generative methods have been proposed to generate real-world data, such as variational auto-encoders (VAE)~\cite{kingma2013auto} and generative adversarial networks (GAN)~\cite{goodfellow2014generative}. However, both VAE and GAN are not able to generate node attributes that are associated with complex and irregular graphs. The adversarial learning idea from GAN has been developed as many other generative approaches. For example, UNIT~\cite{liu2017unsupervised} makes unsupervised image-to-image translation based on the shared-latent space assumption where all images come from the same latent space. However, simply extending this method to the graph domain is infeasible because: (1) the graph structures and node attributes come from two heterogeneous spaces; and (2) the node attributes can either be real-valued or categorical, which makes adversarial learning in data space infeasible.

To this end, we propose a latent-space adversarial learning based method to generate node attributes; called node attribute neural generator (NANG). The proposed NANG is compatible to generate both real-valued and categorical node attributes. Specifically, we assume that the attribute and structure information come from the same latent factor, which can be translated to both attribute modality and structure modality; we call it~\emph{the shared latent factor assumption}. We then use this latent factor as a bridge to convert information from one modality to another, and further generate node attributes based on graph structures.
The main contributions of our work can be summarized as follows:
\begin{itemize}
\setlength{\itemsep}{0pt}
\setlength{\parsep}{0pt}
\setlength{\parskip}{0pt}
 \item Under the shared latent factor assumption, we propose novel node attribute neural generator (NANG), which is compatible to both real-valued and categorical attributes;
 \item We introduce practical measures to evaluate the quality of generated node attributes;
 \item Empirical results on  four  real-world datasets show that our method can handle the node classification and profiling task for those nodes whose attributes are completely inaccessible.
\end{itemize}

\section{Related Work}
\subsection{Deep Generative Models}
Deep generative models have attracted a lot of attention recently. One of the earliest deep generative model is wake-sleep algorithm~\cite{hinton1995wake} which trains the wake phase denoted as the generative model and the sleep phase denoted as the inference model, separately. VAE~\cite{kingma2013auto} and its variants~\cite{Zheng2018DegenerationIV,Zheng2019UnderstandingVI,zhao2017infovae} are proposed to learn the generative model and inference model jointly by maximizing the variational lower bound with reparameterization tricks.

In recent years, GAN~\cite{goodfellow2014generative}, as another family of deep generative models~\cite{hu2017unifying}, has emerged as a hot research topic. It contains a generator and a discriminator, where the discriminator tries to distinguish the real samples and the fake samples and the generator tries to confuse the discriminator. Several works~\cite{goodfellow2014generative,nowozin2016f,chen2016infogan} have pointed that the adversarial loss in GAN actually minimizes a lower bound on Jensen-Shannon divergence (JSD) between the data distribution and the generator distribution. Original GAN has risk facing the vanishing gradient and mode collapse problem. To handle this, a lot of works have been developed to improve the objective function such as f-GAN~\cite{nowozin2016f}, LSGAN~\cite{mao2017least} and WGAN~\cite{arjovsky2017wasserstein}. 
The idea of adversarial learning from GAN is not only limited to generation setting but also can be applied to many other applications such as domain adaptation~\cite{ganin2016domain} and domain translation~\cite{isola2017image,liu2017unsupervised}.

\subsection{Deep Representation Learning on Graphs} 
With the great success of deep neural networks in modeling speech signals, images and text contents, many researchers start to apply deep neural networks to the graph domain. For example, DeepWalk~\cite{perozzi2014deepwalk} and Node2Vec~\cite{grover2016node2vec} learn node embedding by random walks and skip-gram~\cite{mikolov2013efficient} model. Graph convolution networks (GCN) is proposed in~\cite{defferrard2016convolutional,kipf2016semi} and it is successfully applied in semi-supervised node classification problem. A hierarchical and differentiable pooling is proposed in~\cite{ying2018hierarchical} to learn latent representation for graphs. GraphSAGE~\cite{hamilton2017inductive} introduces neighbour sampling and different aggregation manners to make inductive graph convolution on large graphs.

In many real-world applications, graph structures may be missing, incomplete and inaccessible. To handle this issue, graph structure generation has been studied recently. A junction tree variational auto-encoder is proposed in~\cite{jin2018junction} to generate molecular graphs. MolGAN~\cite{de2018molgan} is an implicit and likelihood-free generative model to generate small molecular graphs. GraphRNN~\cite{you2018graphrnn} and NetGAN~\cite{bojchevski2018netgan} generate realistic graphs by the combination of random walks and RNNs. Despite the great potential in many applications, few works consider generating node attributes associated with graphs.

\section{Node Attribute Neural Generator (NANG)}
\subsection{Problem Statement}
Let $\mathcal{G}=(\mathcal{V},A, X)$ be a graph with node set $\mathcal{V}=\{v_{1},v_{2},...,v_{N}\}$, $A\in R^{N\times N}$ be the graph adjacent matrix and $X\in R^{N\times F}$ be the node attribute matrix. Note that the element in $X$ could has either categorical value or real value. 
Let $\mathcal{V}_{o}$ be the set of nodes that are associated with observed attributes. The corresponding node attributes matrix is $X_{o}\in R^{N_{o}\times F}$. Let $\mathcal{V}_{u}$ be the set of nodes that are associated with unobserved attributes. The corresponding node attributes matrix is $X_{u}\in R^{N_{u}\times F}$, which is unknown. To clarify more clearly, we have $V=\mathcal{V}_{u} \cup \mathcal{V}_{o}$, $\mathcal{V}_{u} \cap \mathcal{V}_{o} = \emptyset$ and $N = N_{o} + N_{u}$. Our task aims to generate the unobserved node attributes $X_{u}$ based on the observed node attributes $X_{o}$ and the graph adjacent matrix $A$.

\subsection{Model Formulation}
Graph-structured data include two aspects: node attributes and graph structures. We assume that they share the same latent factor $Z$, which can be translated to either the node attributes or graph structures. Based on this, the marginal distributions of attributes $X$ and structures $A$ satisfy:
\begin{equation}
\label{eq:true}
 p(X)=\int p(X|Z) p(Z)dZ,~p(A)= \int p(A|Z)p(Z)dZ,
\end{equation}
where the conditional distributions $p(X|Z)$ and $p(A|Z)$ are referred to the probabilistic decoders that decode $Z$ to $X$ and $A$, respectively. And $p(Z)$ is the prior distribution for latent factor $Z$. 
In order to better utilize the prior $p(Z)$ to guide the learning process, we learn an aggregated distribution $q(Z)$ for the posteriors $q(Z|X)$ and $p(Z|A)$ based on the observed data. Minimizing the distance between $q(Z)$ and $p(Z)$ can encourage the shared latent factor to match the whole distribution of $p(Z)$. 

Therefore, our goal is to learn the aggregated distribution $q(Z)$ and the probabilistic decoders $p(X|Z)$, $p(A|Z)$ based on the partial observed node attributes $X_{o}$ and graph structure $A$. In the attribute generation stage, we use the structure information to generate the missing attributes, more specifically, we infer the attribute information for the attribute missing nodes by encoding their structure information $A$ into latent factor $Z_{A_{u}} \sim q(Z)$, then the factor will be feed into the probabilistic decoder $p(X|Z)$ to get the prediction of their unobserved attributes.
Our NANG achieves this goal in a way of distribution matching~\cite{makhzani2015adversarial,zhao2017infovae}.  In the following parts, we will give the details about distribution matching, followed by the objective function and the implementation

\subsubsection{Inference via Distribution Matching}
\begin{figure*}[t]
\centering
\includegraphics[width=14cm]{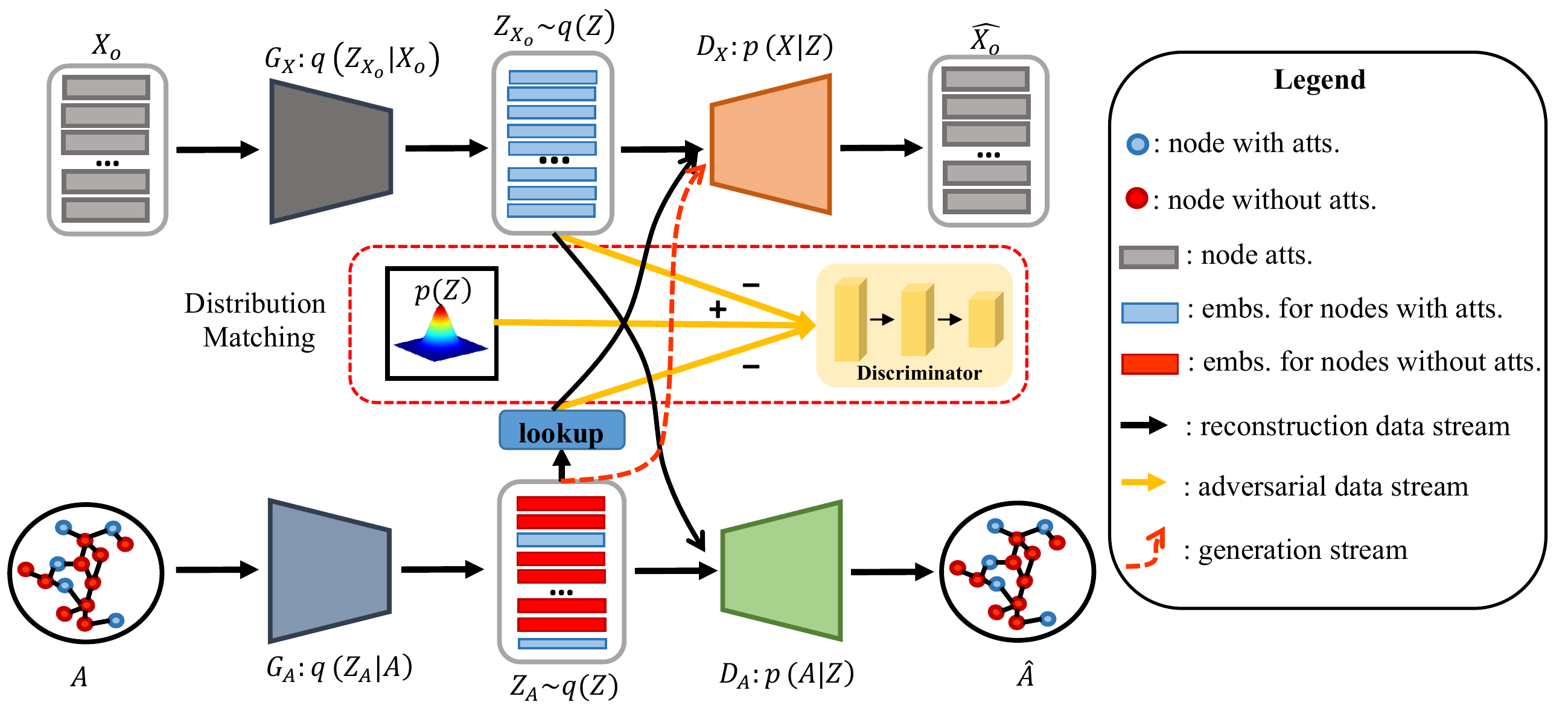}
\caption{Model architecture of NANG. NANG first transforms the node attributes and the graph structures into the latent space, then aligns the latent representation via distribution matching, and finally decodes to the original attributes and structures. In this procedure, information captured from graph structure can be used to recover the node attributes by cross data stream after training.}
\label{figure:model_architecture}
\end{figure*}

NANG learns $q(Z)$ in a way of distribution matching. Specifically, two parameterized encoders $q(Z_{X_{o}}|X_{o})$ and $q(Z_{A}|A)$ are used to encode the attribute information and structure information, respectively. Then we introduce adversarial learning to encourage distribution matching between $q(Z_{X_{o}}|X_{o})$ and $q(Z_{A}|A)$ with prior $p(Z)$. In this way, the latent variable $Z_{X_{o}}$ and $Z_{A}$ are supposed to come from the same aggregated posterior distribution $q(Z)$~\cite{makhzani2015adversarial,makhzani2018implicit}. Besides, the distance between $q(Z)$ and $p(Z)$ is also minimized, encouraging the shared latent factor to match the whole distribution of $p(Z)$.
Let $Z_{A}$ be the latent codes encoded from structure information for all nodes and $A_{o}$ represent the structure information for attribute-observed nodes. Let $Z_{X_{o}}$ and $Z_{A_{o}}$ denote the latent codes respectively encoded from attributes and structures for attribute-observed nodes. To keep consistency with our shared latent factor assumption for decoders $p(X|Z)$ and $p(A|Z)$, they need to reconstruct $Z_{X_{o}}$ (\textit{resp.} $Z_{A}$) to $X_{o}$ (\textit{resp.} $A$), and translate $Z_{A_{o}}$ (\textit{resp.} $Z_{X_{o}}$) to $X_{o}$ (\textit{resp.} $A_{o}$). The architecture is shown in Figure~\ref{figure:model_architecture}. The objective function is formulated as:
\begin{equation}
\label{eq:objective}
\begin{split}
 \min_{\Theta}\max_{D} \mathcal{L} =~& -\mathbb{E}_{X_{o}\sim p_{X_{o}}}[\mathbb{E}_{q(Z_{X_{o}}|X_{o})}[\log p(X_{o}|Z_{X_{o}})]]-\mathbb{E}_{A\sim p_{A}}[\mathbb{E}_{q(Z_{A}|A)}[\log p(A|Z_{A})]] \\
 &-\lambda_\mathrm{c} \cdot \mathbb{E}_{A\sim p_{A}}[\mathbb{E}_{q(Z_{A}|A)}[\log p(X_{o}|Z_{A_{o}})]]-\lambda_\mathrm{c}\cdot \mathbb{E}_{X_{o}\sim p_{X_{o}}} [\mathbb{E}_{q(Z_{X_{o}}|X_{o})}[\log p(A_{o}|Z_{X_{o}})]] \\
 & -\mathcal{L}_{{adv}_{1}}(Z_{T},Z_{X_{o}};G_{X},D)-\mathcal{L}_{{adv}_{2}}(Z_{T},Z_{A_{o}};G_{A},D)
\end{split}
\end{equation}
where $\Theta$ is the network parameters including the encoders (generators) $G_{X}:q(Z_{X_{o}})$ and $G_{A}:q(Z_{A}|A)$ and the decoders $D_{X}:p(X|Z)$ and $D_{A}:p(A|Z)$.
And $D$ is the shared discriminator for adversarial learning $\mathcal{L}_{{adv}_{1}}$ and $\mathcal{L}_{{adv}_{2}}$. $Z_{T}$ indicates true samples we sampled from the prior distribution $p(Z)$ for adversarial learning (we apply standard Gaussian prior here). $\lambda_\mathrm{c}>1.0$ is a hyperparameter to emphasize the second two terms. The first two terms represent the self-reconstruction stream, which means information from attribute to attribute and from structure to structure. The second two terms represent the cross-reconstruction stream, which means information from structure to attribute and from attribute to structure. The last two terms are the adversarial loss defined in Eq.~\ref{eq:adversarial_loss}, which naturally introduce regularization to the objective function.
\begin{equation}
\label{eq:adversarial_loss}
\begin{split}
 & \mathcal{L}_{{adv}_{1}}(Z_{T},Z_{X_{o}};G_{X},D)=-\mathbb{E}_{Z_{T}\sim p(Z)}[\log D(Z_{T})]-\mathbb{E}_{Z_{X_{o}}\sim G_{X}(X_{o})}[\log (1-D(Z_{X_{o}}))], \\
 & \mathcal{L}_{{adv}_{2}}(Z_{T},Z_{A_{o}};G_{A},D)=-\mathbb{E}_{Z_{T}\sim p(Z)}[\log D(Z_{T})]-\mathbb{E}_{Z_{A_{o}}\sim G_{A}(A)}[\log (1-D(Z_{A_{o}}))],
\end{split}
\end{equation}

For the objective function in Eq.~\ref{eq:objective}, the self-reconstruction stream and the cross-reconstruction stream together with our adversarial learning indicate a coupling mechanism which makes information supplement and restores the original non-independent characteristic of node attributes and structures. In this way, the aggregated distribution $q(Z)$, probabilistic decoders $p(X|Z)$ and $p(A|Z)$ can be well captured and we can generate the unobserved node attributes $X_{u}$ through cross reconstruction stream. 

\subsubsection{Implementation}
We formulate our method NANG in an adversarial learning manner that encourages $q(Z)$ learned from observed data to match the whole distribution of $p(Z)$, the method architecture is shown in Figure~\ref{figure:model_architecture}. NANG consists of three modules: (1) self-reconstruction stream, (2) cross-reconstruction stream and (3) adversarial regularization.

In the self-reconstruction stream, both the observed attributes $X_{o}$ and structural $A$ are encoded into the latent space by two encoders $G_{X}$ and $G_{A}$. The corresponding latent codes $Z_{X_{o}}$ and $Z_{A}$ are decoded as $X_{o}$ and $A$ by two decoders $D_{X}$ and $D_{A}$. This appears in Figure~\ref{figure:model_architecture}, in which $G_{X}$ is a two-layer MLP, $G_{A}$ is a two-layer GCN, $D_{X}$ is a two-layer MLP and $D_{A}$ is a two-layer MLP followed by $\sigma(x\cdot x^{T})$ ($x$ represents the output of two-layer MLP in $D_{A}$ and $\sigma$ denotes the sigmoid function).
Let $A_{o}$ be the structure information for attribute-observed nodes and $Z_{A_{o}}$ be the corresponding latent codes encoded by $G_{A}$. In the cross-reconstruction stream, the latent codes $Z_{X_{o}}$ from $G_{X}$ is decoded as $A_{o}$ by $D_{A}$ and the latent codes $Z_{A_{o}}$ from $G_{A}$ is decoded as $X_{o}$ by $D_{X}$. 
In the adversarial regularization module, we apply adversarial learning between $Z_{X_{o}}$, $Z_{A_{o}}$ and samples from standard Gaussian prior $p(Z)$, sharing the same two-layer MLP discriminator network. These three modules work together and encourage the learning process of $q(Z)$, $p_{X}(X|Z)$ and $p_{A}(A|Z)$. 

\section{Experiments}
\subsection{Experimental Setup}
\textbf{Datasets.} We evaluate the proposed NANG on four real-world datasets to quantify the performance.
\begin{itemize}
\setlength{\itemsep}{0pt}
\setlength{\parsep}{0pt}
\setlength{\parskip}{0pt}
 \item Cora. Cora~\cite{mccallum2000automating} a citation graph whose nodes are papers and edges are citation links. It contains 10,556 edges and 2,708 papers that are categorized into 7 classes. Each node attribute vector indicates whether the corresponding paper contains certain word tokens and it is represented as a multi-hot vector with dimension 1,433.
 \item Citeseer. Citeseer~\cite{sen2008collective} is also a citation graph which contains 9,228 edges and 3,327 papers that are categorized into 6 classes. After stemming and stop word removal operation of the content, 3,704 distinct words make up the attribute corpus. Each node attribute vector is formed from the corpus and represented as a multi-hot vector with dimension 3,704.
 \item Steam. Steam\footnote{https://store.steampowered.com/} is a dataset we collected from a game official website with the user-bought behavior history of 9,944 items and 352 labels for these items.
 We count the co-purchase frequency between every two games and make a sparse item co-purchase graph through binarilization operation with the threshold as 10. After that, we obtain 533,962 edges for this graph. The label corpus constructs the multi-hot attribute vector for each item with dimension 352.
 \item Pubmed. Pubmed~\cite{namata2012query} is a citation graph where nodes are categorized into 3 classes. There are 19,717 nodes and 88,651 edges in this graph. Each node attribute vector is described by a Term Frequency-Inverse Document Frequency (TF-IDF) vector from 500 distinct terms.
\end{itemize}
Among these datasets, attributes of Cora, Citeseer and Steam are categorical and represented as multi-hot vectors. For Pubmed, node attributes are real valued and represented as scalars. The statistics of these datasets is illustrated in Appendix~\ref{appendix:experimental_setup} Table~\ref{table:dataset}.

\textbf{Baselines of Generation Methods.} To evaluate the generation performance of our method, we compare it with three baselines.
\begin{itemize}
\setlength{\itemsep}{0pt}
\setlength{\parsep}{0pt}
\setlength{\parskip}{0pt}
 \item NeighAggre. NeighAggre aims to directly aggregate the neighbors' attributes for nodes without attributes through mean pooling. When a neighbor' attributes are missing, we do not regard it as the aggregation node.
 \item VAE. Although VAE~\cite{kingma2013auto} does not naturally suit this task, we consider it as a baseline with some tricks. We make normal VAE for the attribute-observed nodes, encoding attributes of these nodes as latent codes. For the nodes without any attribute, we use neighbour aggregation like NeighAggre in the latent space. Then, the decoder in VAE can be used to generate node attributes.
 \item GCN. For GCN~\cite{kipf2016semi} as a baseline, only the structure information is used as input and encoded as latent embeddings. Then the latent embeddings are decoded by an additional two-layer MLP, being supervised by the observed attributes. In the test stage, we use the latent embeddings of test nodes to generate node attributes through the two-layer MLP decoder.
\end{itemize}

\subsection{Applications and Evaluation Measures}
Generative models such as VAE based~\cite{kingma2013auto,higgins2017beta} and GAN based~\cite{isola2017image,liu2017unsupervised} strike in generating real-like data sharing the same distribution as true data. They mainly use reconstruction error or visual effect to evaluate the generation performance. Especially for GAN based generative methods, other objective functions such as MMD, total variance and Wasserstein distance are proposed to measure the distance between true data distribution $p(X)$ and generated data distribution $p_{g}(X)$~\cite{hu2017unifying}. However, in our problem, whether the generated attributes can benefit real-world applications is more considerable. Consequently, we propose to measure the quality of generated node attributes from both node level and attribute level with two real-world applications.
\begin{itemize}
\setlength{\itemsep}{0pt}
\setlength{\parsep}{0pt}
\setlength{\parskip}{0pt}
 \item Node classification. This node classification task aims to criticize whether the generated node attributes serve as data augmentation and benefit the classification model. In this task, we use the generated attributes of test nodes to make node classification comparison among different methods, taking accuracy as the evaluation metric. In other words, this task evaluates the overall quality of generated node attributes by classification methods, which is also termed as node level evaluation. We implement this task on Cora, Citeseer and Pubmed since they have node class information.
 \item Profiling. Profile provides cognitive description for objects such as key terms for  papers on Cora, Citeseer and labels for items on Steam. Profiling aims to predict the possible profile for test nodes, we use Recall@k and NDCG@k as the evaluation metrics. In other words, this task evaluates the recall and ranking of generated node attributes, which is also termed as attribute level evaluation. For this task, we compare different methods on Cora, Citeseer and Steam since attributes of these datasets are categorical.
\end{itemize}


\begin{table}[t]
\centering
\caption{Evaluation of generated attributes on node classification task. The first column with "X", "A" and "A+X" indicates three settings to do node classification, with only attributes, only structures and the fused one, respectively. NANG-Cross means only cross-reconstruction loss and GAN loss are applied. Similarly, NANG-Self represents only self-reconstruction loss and GAN loss are used. True atts. represents we use the ground truth attributes to do node classification.}
\label{table:classification}
\begin{tabular}{c|ccccc}
\hline
 & atts. generation method & classification method & Cora & Citeseer & Pubmed \\ \hline
\multirow{7}{*}{X} & NeighAggre & MLP & 0.6248 & 0.5539 & \textbf{0.5150} \\
 & VAE & MLP & 0.2826 & 0.2551 & 0.4008 \\
 & GCN & MLP & 0.3943 & 0.3768 & 0.3992 \\ \cline{2-6} 
 & NANG-Cross & MLP & 0.7074 & 0.4976 & 0.4000 \\
 & NANG-Self & MLP & 0.3036 & 0.2289 & 0.4023 \\
 & \textbf{NANG} & \textbf{MLP} & \textbf{0.7644} & \textbf{0.6010} &0.4652 \\ \cline{2-6} 
 & True atts. & MLP & 0.7618 & 0.7174 & 0.656 \\ \hline
\multirow{3}{*}{A} & - & DeepWalk+MLP & 0.7149 & 0.4802 & 0.6917 \\
 & - & Node2Vec+MLP & 0.6830 & 0.4422 & 0.6721 \\
 & - & GCN & 0.7631 & 0.5651 & 0.7125 \\ \hline
\multirow{7}{*}{A+X} & NeighAggre & GCN & 0.6494 & 0.5413 & 0.6564 \\
 & VAE & GCN & 0.3011 & 0.2663 & 0.4007 \\
 & GCN & GCN & 0.4387 & 0.4079 & 0.4203 \\ \cline{2-6} 
 & NANG-Cross & GCN & 0.7727 & 0.5358 & 0.4197 \\
 & NANG-Self & GCN & 0.3402 & 0.2698 & 0.4204 \\
 & \textbf{NANG} & \textbf{GCN} & \textbf{0.8327} & \textbf{0.6599} & \textbf{0.7537} \\ \cline{2-6} 
 & True atts. & GCN & 0.8493 & 0.7348 & 0.8723 \\ \hline
\end{tabular}
\end{table}

\subsection{Node Classification}
\label{sec:node_classification}
In the task of node classification, the generated node attributes are split into 80\% train data and 20\% test data with five-fold validation performed 10 times. We consider two classifiers, including MLP and GCN which both use the class information as supervision. Three settings are designed to conduct the comparisons: the node-attribute-only approach, the graph-structure-only approach, and the fused approach. In the node-attribute-only approach, we directly use the generated attributes and a two-layer MLP as the classifier to do the classification task; in the graph-structure-only approach, we only use the graph structure without considering the node attributes, which has been studied by many methods such as DeepWalk~\cite{perozzi2014deepwalk}, Node2Vec~\cite{grover2016node2vec} and GCN~\cite{kipf2016semi}. DeepWalk and Node2Vec both aim to learn node embeddings and then a MLP classifier is used. While GCN is an end-to-end method which learns the node embeddings supervised by the classification loss. In the fused approach, we combine the generated node attributes and structure information with GCN classifier.

Table~\ref{table:classification} shows the classification performance, in which "X" indicates that only generated node attributes are used, "A" indicates that only structural information is used and "A+X" is the fused one. We can summarize that: (1) When only using the generated attributes to do node classification task, the proposed NANG can obtain significant gain over baseline methods: NeighAggre, VAE and GCN. Compared to the most competitive method NeighAggre, the proposed NANG reaches nearly 14\% and 5\% gain on Cora and Citeseer, respectively. On Pubmed, it seems that NeighAggre suits this dataset and this setting well, but it deteriorates quickly when less attribute-observed nodes exist, which will be shown in Section~\ref{section:less_nodes}. (2) The performance of NANG gets closer to that of true attributes. NANG even gets better than the true attributes on Cora mainly because the generated attributes of NANG may contain some graph structure information that is beneficial to the classification task. (3) Both NANG-Cross and NANG-Self perform worse than NANG, because the incompleteness of our objective function cannot guarantee the shared latent factor assumption.

Apart from the generated attributes, we can also use graph structure information to make node classification and the result is shown in the Table~\ref{table:classification} with the "A" signed row. Among these methods, GCN performs better, which is in accordance with recent works. This also inspires us that whether the generated attributes of our method could augment the GCN classification performance. Therefore, we conduct the fused "A+X" experiment. From the comparison between the "A" row and the "A+X" row in Table~\ref{table:classification}, we can summarize that: (1) The generated attributes from NANG can augment the GCN classification performance with 6.96\%, 9.48\% and 4.12\% gain on Cora, Citeseer and Pubmed, respectively. While NeighAggre fails and harms the GCN performance with the figure as 11.37\%, 2.38\% and 5.61\% on Cora, Citeseer and Pubmed, respectively. (2) The generated attributes of other methods such as GCN and VAE are of inferior quality and they hurt the GCN performance a lot, because they cannot capture the complex translation pattern between attribute and structure modality.

\begin{table}[t]
\small
\centering
\caption{Evaluation for generated attributes on profiling task. Note that the average non-zero attribute number for Cora and Citeseer is 18.17 and 31.6, respectively. Therefore, we use top 10, 20, 50 to evaluate the performance on these two datasets. Similarly, we use top 3, 5, 10 to evaluate the performance on Steam.}
\label{table:label_propagation}
\begin{tabular}{ccccccc}
\hline
\multicolumn{7}{c}{\textbf{Cora}}   \\ \hline
\multicolumn{1}{c|}{Method} & Recall@10 & Recall@20 & Recall@50 & NDCG@10 & NDCG@20 & NDCG@50 \\ \hline
\multicolumn{1}{c|}{NeighAggre} & 0.0906 & 0.1413 & 0.1961 & 0.1217 & 0.1548 & 0.1850 \\
\multicolumn{1}{c|}{VAE} & 0.0887 & 0.1228 & 0.2116 & 0.1224 & 0.1452 & 0.1924 \\
\multicolumn{1}{c|}{GCN} & 0.1271 & 0.1772 & 0.2962 & 0.1736 & 0.2076 & 0.2702 \\ \hline
\multicolumn{1}{c|}{NANG-Cross} & 0.1378 & 0.2018 & 0.3339 & 0.1931 & 0.2360 & 0.3052 \\
\multicolumn{1}{c|}{NANG-Self} & 0.1224 & 0.1724 & 0.2823 & 0.1686 & 0.2023 & 0.2599 \\
\multicolumn{1}{c|}{\textbf{NANG}} & \textbf{0.1508} & \textbf{0.2182} & \textbf{0.3429} & \textbf{0.2112} & \textbf{0.2546} & \textbf{0.3212} \\ \hline
\multicolumn{7}{c}{\textbf{Citeseer}}    \\ \hline
\multicolumn{1}{c|}{Method} & Recall@10 & Recall@20 & Recall@50 & NDCG@10 & NDCG@20 & NDCG@50 \\ \hline
\multicolumn{1}{c|}{NeighAggre} & 0.0511 & 0.0908 & 0.1501 & 0.0823 & 0.1155 & 0.1560 \\
\multicolumn{1}{c|}{VAE} & 0.0382 & 0.0668 & 0.1296 & 0.0601 & 0.0839 & 0.1251 \\
\multicolumn{1}{c|}{GCN} & 0.0620 & 0.1097 & 0.2052 & 0.1026 & 0.1423 & 0.2049 \\ \hline
\multicolumn{1}{c|}{NANG-Cross} & 0.0679 & 0.1163 & 0.2140 & 0.1167 & 0.1570 & 0.2209 \\
\multicolumn{1}{c|}{NANG-Self} & 0.0564 & 0.1013 & 0.1963 & 0.0863 & 0.1238 & 0.1860 \\
\multicolumn{1}{c|}{\textbf{NANG}} & \textbf{0.0764} & \textbf{0.1280} & \textbf{0.2377} & \textbf{0.1298} & \textbf{0.1729} & \textbf{0.2447} \\ \hline
\multicolumn{7}{c}{\textbf{Steam}}    \\ \hline
\multicolumn{1}{c|}{Method} & Recall@3 & Recall@5 & Recall@10 & NDCG@3 & NDCG@5 & NDCG@10 \\ \hline
\multicolumn{1}{c|}{NeighAggre} & 0.0603 & 0.0881 & 0.1446 & 0.0955 & 0.1204 & 0.1620 \\
\multicolumn{1}{c|}{VAE} & 0.0564 & 0.0820 & 0.1251 & 0.0902 & 0.1133 & 0.1437 \\
\multicolumn{1}{c|}{GCN} & 0.2392 & 0.3258 & 0.4575 & 0.3366 & 0.4025 & 0.4848 \\ \hline
\multicolumn{1}{c|}{NANG-Cross} & 0.2429 & 0.3116 & 0.4614 & 0.3414 & 0.3969 & 0.4889 \\
\multicolumn{1}{c|}{NANG-Self} & 0.2382 & 0.3381 & 0.4611 & 0.3282 & 0.4057 & 0.4835 \\
\multicolumn{1}{c|}{\textbf{NANG}} & \textbf{0.2527} & \textbf{0.3560} & \textbf{0.4933} & \textbf{0.3544} & \textbf{0.4332} & \textbf{0.5215} \\ \hline
\end{tabular}
\end{table}

\subsection{Profiling}
For this profiling task, the generated attributes on Cora, Citeseer and Steam are probabilities that the node may have in each attribute dimension. Good generated attributes should have high probability in specific attribute dimension as the true attributes. Accordingly, taking the recall ability and ranking ability into consideration, we use Recall@k and NDCG@k to evaluate the attribute generation performance in attribute level. The result is shown in Table~\ref{table:label_propagation}.

From this table, it is clear that NeighAggre performs the worst among all methods on the three datasets, especially for Steam since it is not a learning algorithm and cannot generate reliable attributes for profiling in attribute level. However, NANG generates attributes based on the translation knowledge from structure information to attribute information, which is more adaptable and flexible. Indeed, results in Table~\ref{table:label_propagation} show NANG achieves superior performance over other methods for profiling on this attribute level evaluation. Compared to GCN, NANG reaches a 4.67\% and 3.25\% gain of Recall@50 on Cora and Citeseer, respectively.


\begin{figure*}[ht]
\centering
\begin{minipage}[t]{0.32\textwidth}
\centering
\includegraphics[width=\textwidth]{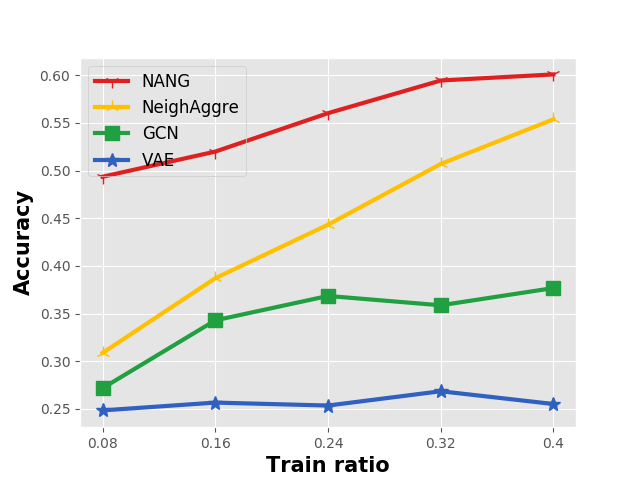}
\vspace{-15pt}
\caption*{(a) X - Citeseer}
\end{minipage}
\begin{minipage}[t]{0.32\textwidth}
\centering
\includegraphics[width=\textwidth]{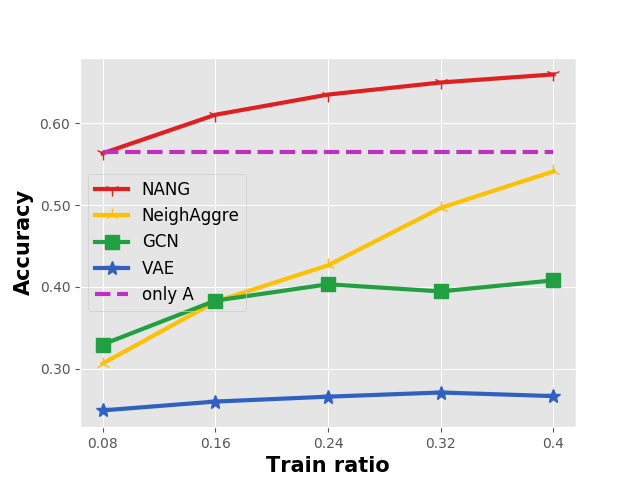}
\vspace{-15pt}
\caption*{(b) A+X - Citeseer}
\end{minipage}
\begin{minipage}[t]{0.32\textwidth}
\centering
\includegraphics[width=\textwidth]{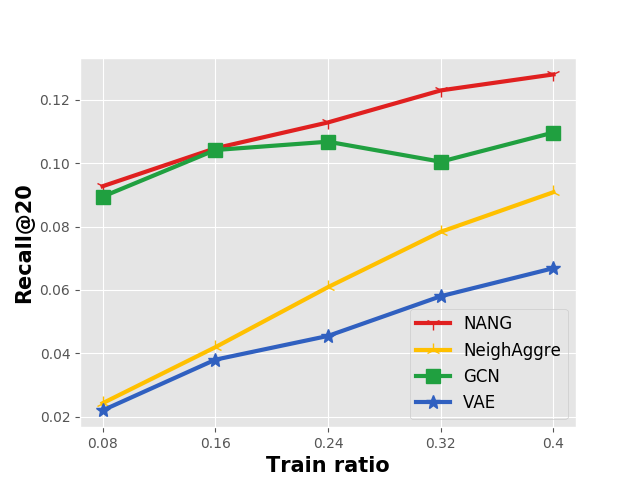}
\vspace{-15pt}
\caption*{(c) Recall - Citeseer}
\end{minipage} \\
\begin{minipage}[t]{0.32\textwidth}
\centering
\includegraphics[width=\textwidth]{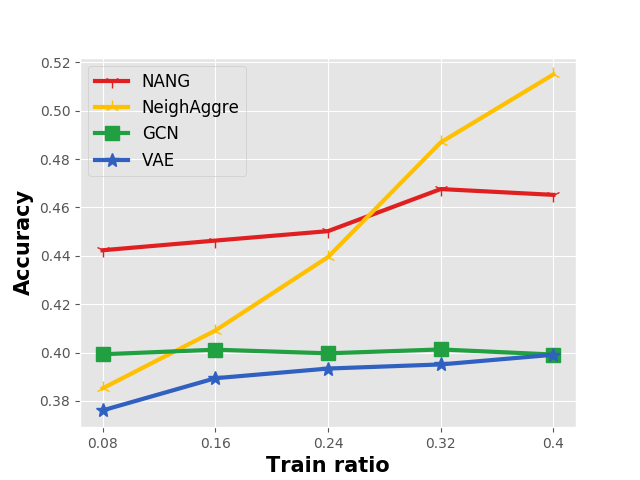}
\vspace{-15pt}
\caption*{(d) X - Pubmed}
\end{minipage}
\begin{minipage}[t]{0.32\textwidth}
\centering
\includegraphics[width=\textwidth]{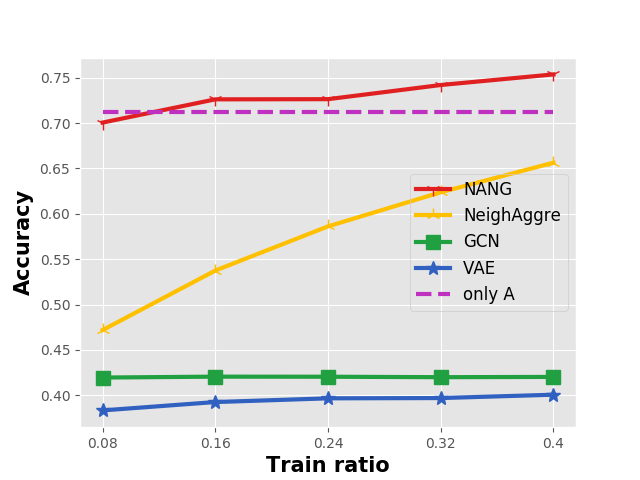}
\vspace{-15pt}
\caption*{(e) A+X - Pubmed}
\end{minipage}
\begin{minipage}[t]{0.32\textwidth}
\centering
\includegraphics[width=\textwidth]{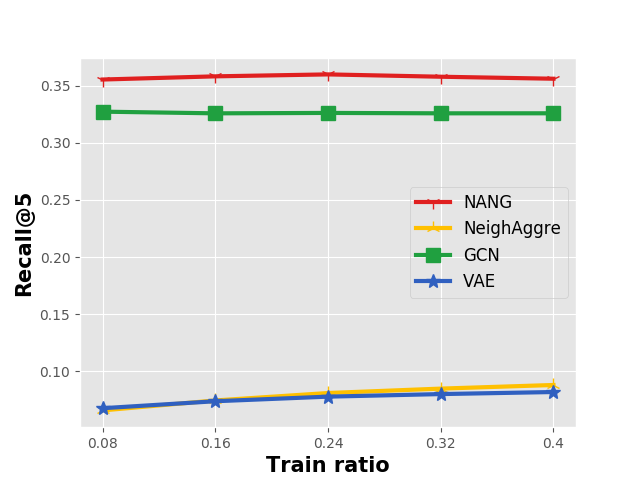}
\vspace{-15pt}
\caption*{(f) Recall - Steam}
\end{minipage}
\caption{Node classification and profiling performance with different ratio of training nodes. (a)(d) illustrate the result for node classification with "X" setting on Citeseer and Pubmed, respectively. (b)(e) show the result for node classification with "A+X" setting on Citeseer and Pubmed, respectively. The dotted line is a criterion to criticize whether the generated attributes can enhance the GCN classifier. In other words, it represents only structure information "A" is used by GCN to do the classification task. (c)(f) show the result for profiling on Citeseer and Steam, respectively.}
\label{figure:fts_sparsity_res}
\end{figure*}

\subsection{Less Attribute-observed Nodes}
\label{section:less_nodes}
The partial attribute-observed nodes are necessary and supervise this node attribute generation task. In some scenarios, this supervised information could be less, so it is necessary to see whether our NANG can still generate reliable and high-quality node attributes when less attributed nodes are observed. We conduct an experiment to explore the node classification and profiling performance under this condition. The result is shown in Figure~\ref{figure:fts_sparsity_res}.

In Figure~\ref{figure:fts_sparsity_res} (a)(d) for node classification when only "X" is used, we can see that NANG performs much better than other methods on Citeseer. The gap is more obvious when the attribute-observed nodes are less. On Pubmed, although NeighAggre performs better than NANG when attributes-observed nodes are more, it is not robust when the attributes-observed nodes are less. Figure~\ref{figure:fts_sparsity_res} (b)(e) demonstrate the node classification performance when "A+X" is used. In these two figures, the dotted line represents only "A" is used by a GCN classifier, which is denoted as a criterion to criticize whether the generated attributes could enhance the GCN classifier. It is clear that, in the "A+X" setting, our NANG reaches superior performance than other methods. Besides, it is robust when the attribute-observed nodes are less while NeighAggre fails to enhance the GCN classifier. Figure~\ref{figure:fts_sparsity_res} (c)(f) show the profiling performance on Citeseer and Steam. It is clear that NANG can perform better than other methods when attribute-observed nodes are less, because NANG learns the translation knowledge from attribute to structure modality which is more adaptive and more flexible.

\section{Conclusions}
In this paper, we propose an adversarial learning method called NANG for the node attribute generation problem. In NANG, we assume that both node attribute information and structure information share the same latent factor which can be translated as different modalities. The implicit distribution for this latent factor is modeled in an auto-encoding Bayes and adversarial learning manner. We further evaluate the quality of generated node attributes on both node level and attribute level through practical applications. Empirical results validate the superiority of our method on node classification and profiling for those nodes whose attributes are completely inaccessible.

\newpage
\bibliographystyle{plain}\small
\bibliography{nips_2019}

\begin{thebibliography}{10}

\bibitem{arjovsky2017wasserstein}
Martin Arjovsky, Soumith Chintala, and L{\'e}on Bottou.
\newblock Wasserstein generative adversarial networks.
\newblock In {\em International Conference on Machine Learning}, pages
  214--223, 2017.

\bibitem{bojchevski2018netgan}
Aleksandar Bojchevski, Oleksandr Shchur, Daniel Z{\"u}gner, and Stephan
  G{\"u}nnemann.
\newblock Netgan: Generating graphs via random walks.
\newblock In {\em Proceedings of the 35th International Conference on Machine
  Learning}, volume~80 of {\em Proceedings of Machine Learning Research}, pages
  610--619, Stockholmsmässan, Stockholm Sweden, 10--15 Jul 2018. PMLR.

\bibitem{chen2016infogan}
Xi~Chen, Yan Duan, Rein Houthooft, John Schulman, Ilya Sutskever, and Pieter
  Abbeel.
\newblock Infogan: Interpretable representation learning by information
  maximizing generative adversarial nets.
\newblock In {\em Advances in neural information processing systems}, pages
  2172--2180, 2016.

\bibitem{de2018molgan}
Nicola De~Cao and Thomas Kipf.
\newblock {MolGAN: An implicit generative model for small molecular graphs}.
\newblock In {\em ICML 2018 workshop on Theoretical Foundations and
  Applications of Deep Generative Models}, 2018.

\bibitem{defferrard2016convolutional}
Micha{\"e}l Defferrard, Xavier Bresson, and Pierre Vandergheynst.
\newblock Convolutional neural networks on graphs with fast localized spectral
  filtering.
\newblock In {\em Advances in neural information processing systems}, pages
  3844--3852, 2016.

\bibitem{fu2015transductive}
Yanwei Fu, Timothy~M Hospedales, Tao Xiang, and Shaogang Gong.
\newblock Transductive multi-view zero-shot learning.
\newblock {\em IEEE transactions on pattern analysis and machine intelligence},
  37(11):2332--2345, 2015.

\bibitem{ganin2016domain}
Yaroslav Ganin, Evgeniya Ustinova, Hana Ajakan, Pascal Germain, Hugo
  Larochelle, Fran{\c{c}}ois Laviolette, Mario Marchand, and Victor Lempitsky.
\newblock Domain-adversarial training of neural networks.
\newblock {\em The Journal of Machine Learning Research}, 17(1):2096--2030,
  2016.

\bibitem{goodfellow2014generative}
Ian Goodfellow, Jean Pouget-Abadie, Mehdi Mirza, Bing Xu, David Warde-Farley,
  Sherjil Ozair, Aaron Courville, and Yoshua Bengio.
\newblock Generative adversarial nets.
\newblock In {\em Advances in neural information processing systems}, pages
  2672--2680, 2014.

\bibitem{grover2016node2vec}
Aditya Grover and Jure Leskovec.
\newblock node2vec: Scalable feature learning for networks.
\newblock In {\em Proceedings of the 22nd ACM SIGKDD international conference
  on Knowledge discovery and data mining}, pages 855--864. ACM, 2016.

\bibitem{hamilton2017inductive}
Will Hamilton, Zhitao Ying, and Jure Leskovec.
\newblock Inductive representation learning on large graphs.
\newblock In {\em Advances in Neural Information Processing Systems}, pages
  1024--1034, 2017.

\bibitem{higgins2017beta}
Irina Higgins, Loic Matthey, Arka Pal, Christopher Burgess, Xavier Glorot,
  Matthew Botvinick, Shakir Mohamed, and Alexander Lerchner.
\newblock beta-vae: Learning basic visual concepts with a constrained
  variational framework.
\newblock In {\em International Conference on Learning Representations}, 2017.

\bibitem{hinton1995wake}
Geoffrey~E Hinton, Peter Dayan, Brendan~J Frey, and Radford~M Neal.
\newblock The" wake-sleep" algorithm for unsupervised neural networks.
\newblock {\em Science}, 268(5214):1158--1161, 1995.

\bibitem{hu2017unifying}
Zhiting Hu, Zichao Yang, Ruslan Salakhutdinov, and Eric~P. Xing.
\newblock On unifying deep generative models.
\newblock In {\em International Conference on Learning Representations}, 2018.

\bibitem{isola2017image}
Phillip Isola, Jun-Yan Zhu, Tinghui Zhou, and Alexei~A Efros.
\newblock Image-to-image translation with conditional adversarial networks.
\newblock In {\em Proceedings of the IEEE conference on computer vision and
  pattern recognition}, pages 1125--1134, 2017.

\bibitem{jin2018junction}
Wengong Jin, Regina Barzilay, and Tommi Jaakkola.
\newblock Junction tree variational autoencoder for molecular graph generation.
\newblock In {\em Proceedings of the 35th International Conference on Machine
  Learning}, 2018.

\bibitem{kalofolias2014matrix}
Vassilis Kalofolias, Xavier Bresson, Michael Bronstein, and Pierre
  Vandergheynst.
\newblock Matrix completion on graphs.
\newblock {\em arXiv preprint arXiv:1408.1717}, 2014.

\bibitem{kingma2013auto}
Diederik~P Kingma and Max Welling.
\newblock Auto-encoding variational bayes.
\newblock In {\em International Conference on Learning Representations}, 2014.

\bibitem{kipf2016variational}
Thomas~N Kipf and Max Welling.
\newblock Variational graph auto-encoders.
\newblock In {\em NIPS Workshop on Bayesian Deep Learning}, 2016.

\bibitem{kipf2016semi}
Thomas~N. Kipf and Max Welling.
\newblock Semi-supervised classification with graph convolutional networks.
\newblock In {\em International Conference on Learning Representations}, 2017.

\bibitem{li2019actional}
Maosen Li, Siheng Chen, Xu~Chen, Ya~Zhang, Yanfeng Wang, and Qi~Tian.
\newblock Actional-structural graph convolutional networks for skeleton-based
  action recognition.
\newblock In {\em Proceedings of the IEEE conference on computer vision and
  pattern recognition}, 2019.

\bibitem{liu2017unsupervised}
Ming-Yu Liu, Thomas Breuel, and Jan Kautz.
\newblock Unsupervised image-to-image translation networks.
\newblock In {\em Advances in Neural Information Processing Systems}, pages
  700--708, 2017.

\bibitem{maaten2008visualizing}
Laurens van~der Maaten and Geoffrey Hinton.
\newblock Visualizing data using t-sne.
\newblock {\em Journal of machine learning research}, 9(Nov):2579--2605, 2008.

\bibitem{makhzani2018implicit}
Alireza Makhzani.
\newblock Implicit autoencoders.
\newblock {\em arXiv preprint arXiv:1805.09804}, 2018.

\bibitem{makhzani2015adversarial}
Alireza Makhzani, Jonathon Shlens, Navdeep Jaitly, and Ian Goodfellow.
\newblock Adversarial autoencoders.
\newblock In {\em International Conference on Learning Representations}, 2016.

\bibitem{mao2017least}
Xudong Mao, Qing Li, Haoran Xie, Raymond~YK Lau, Zhen Wang, and Stephen
  Paul~Smolley.
\newblock Least squares generative adversarial networks.
\newblock In {\em Proceedings of the IEEE International Conference on Computer
  Vision}, pages 2794--2802, 2017.

\bibitem{mccallum2000automating}
Andrew~Kachites McCallum, Kamal Nigam, Jason Rennie, and Kristie Seymore.
\newblock Automating the construction of internet portals with machine
  learning.
\newblock {\em Information Retrieval}, 3(2):127--163, 2000.

\bibitem{mikolov2013efficient}
Tomas Mikolov, Kai Chen, Greg Corrado, and Jeffrey Dean.
\newblock Efficient estimation of word representations in vector space.
\newblock In {\em International Conference on Learning Representations}, 2013.

\bibitem{namata2012query}
Galileo Namata, Ben London, Lise Getoor, Bert Huang, and UMD EDU.
\newblock Query-driven active surveying for collective classification.
\newblock In {\em 10th International Workshop on Mining and Learning with
  Graphs}, 2012.

\bibitem{nowozin2016f}
Sebastian Nowozin, Botond Cseke, and Ryota Tomioka.
\newblock f-gan: Training generative neural samplers using variational
  divergence minimization.
\newblock In {\em Advances in neural information processing systems}, pages
  271--279, 2016.

\bibitem{perozzi2014deepwalk}
Bryan Perozzi, Rami Al-Rfou, and Steven Skiena.
\newblock Deepwalk: Online learning of social representations.
\newblock In {\em Proceedings of the 20th ACM SIGKDD international conference
  on Knowledge discovery and data mining}, pages 701--710. ACM, 2014.

\bibitem{garcia2017few}
Victor~Garcia Satorras and Joan~Bruna Estrach.
\newblock Few-shot learning with graph neural networks.
\newblock In {\em International Conference on Learning Representations}, 2018.

\bibitem{sen2008collective}
Prithviraj Sen, Galileo Namata, Mustafa Bilgic, Lise Getoor, Brian Galligher,
  and Tina Eliassi-Rad.
\newblock Collective classification in network data.
\newblock {\em AI magazine}, 29(3):93--93, 2008.

\bibitem{ying2018graph}
Rex Ying, Ruining He, Kaifeng Chen, Pong Eksombatchai, William~L Hamilton, and
  Jure Leskovec.
\newblock Graph convolutional neural networks for web-scale recommender
  systems.
\newblock In {\em Proceedings of the 24th ACM SIGKDD International Conference
  on Knowledge Discovery \& Data Mining}, pages 974--983. ACM, 2018.

\bibitem{ying2018hierarchical}
Rex Ying, Jiaxuan You, Christopher Morris, Xiang Ren, William~L. Hamilton, and
  Jure Leskovec.
\newblock Hierarchical graph representation learning with differentiable
  pooling.
\newblock In {\em Advances in neural information processing systems}, pages
  4800--4810, 2018.

\bibitem{you2018graphrnn}
Jiaxuan You, Rex Ying, Xiang Ren, William Hamilton, and Jure Leskovec.
\newblock {G}raph{RNN}: Generating realistic graphs with deep auto-regressive
  models.
\newblock In Jennifer Dy and Andreas Krause, editors, {\em Proceedings of the
  35th International Conference on Machine Learning}, volume~80 of {\em
  Proceedings of Machine Learning Research}, pages 5708--5717,
  Stockholmsmässan, Stockholm Sweden, 10--15 Jul 2018. PMLR.

\bibitem{zhao2017infovae}
Shengjia Zhao, Jiaming Song, and Stefano Ermon.
\newblock Infovae: Information maximizing variational autoencoders.
\newblock In {\em Proceedings of the 33rd Association for the Advancement of
  Artificial Intelligence}, 2019.

\bibitem{Zheng2018DegenerationIV}
Huangjie Zheng, Jiangchao Yao, Ya~Zhang, and Ivor Wai-Hung Tsang.
\newblock Degeneration in vae: in the light of fisher information loss.
\newblock {\em ArXiv}, abs/1802.06677, 2018.

\bibitem{Zheng2019UnderstandingVI}
Huangjie Zheng, Jiangchao Yao, Ya~Zhang, Ivor Wai-Hung Tsang, and Jia Wang.
\newblock Understanding vaes in fisher-shannon plane.
\newblock In {\em Proceedings of the 33rd Association for the Advancement of
  Artificial Intelligence}, 2019.

\end{thebibliography}

\appendix
\clearpage
\section{Supplementary Experimental Setup}
\label{appendix:experimental_setup}

\textbf{Dataset Statistics.} The statistics of used datasets is shown in Table~\ref{table:dataset}.

\begin{table}[ht]
\caption{The statistics of four datasets. In this table, atts form means the attribute style. $\#$avg nnz atts num means the average hot number for nodes. \#class indicates the number of classes of nodes for different datasets.}
\centering
\label{table:dataset}
\begin{tabular}{c|cccc}
\hline
 & Cora & Citeseer & Steam & Pubmed \\ \hline
\#nodes & 2,708 & 3,327 & 9,944 & 19,717 \\
\#edges & 10,556 & 9,228 & 533,962 & 88,651 \\
\#graph sparsity & 0.14\% & 0.08\% & 0.53\% & 0.02\% \\
\#atts dim & 1,433 & 3,703 & 352 & 500 \\
\#avg nnz atts num & 18.17 & 31.6 & 8.45 & - \\
\#class & 7 & 6 & - & 3 \\
atts form & categorical & categorical & categorical & real-valued \\ \hline
\end{tabular}
\end{table}

\textbf{Parameter Setting.} 
For each dataset, we randomly sample $40\%$ nodes with attributes as training data and $10\%$ as validation data and the rest $50\%$ as test data that need to generate attributes. For NeighAggre, we directly use the one-hop neighbours as a node's neighbours. For all learning based methods (VAE, GCN, NANG), we set the latent dimension as 64 with 0.005 as the learning rate. Dropout rate equals 0.5 is utilized and the maximum iteration number is 1,000. Adam optimizer is applied for them to learn the model parameters. Remind that different datasets may have different attribute forms including categorical and real-valued. Therefore, for the datasets with categorical attributes such as Cora, Citeseer and Steam, weighted Binary Cross Entropy loss (BCE) is applied. The weight put on non-zero samples equals $\frac{\#zero~count}{\#non-zero~count}$ calculated from the training node attribute matrix. And for Pubmed with real-valued attributes, Mean Square Error (MSE) loss is used.

For our adversarial NANG, We set the generation step as 2 and discriminator step as 1 for Cora, Citeseer while the generation step as 5 and discriminator step as 1 for Steam and Pubmed. The hyper-parameters is $\lambda_\mathrm{c}=10$ for Cora, Citeseer, Steam while $\lambda_\mathrm{c}=50$ for Pubmed. The experiments are conducted with multi times and the mean value is adopted as the performance. The method is implemented by Pytorch on one computer with one Nvidia TitanX GPU.

\section{Latent Embedding Visualization}
\begin{figure*}[hbt]
\centering
\begin{minipage}[t]{0.32\textwidth}
\centering
\includegraphics[width=\textwidth]{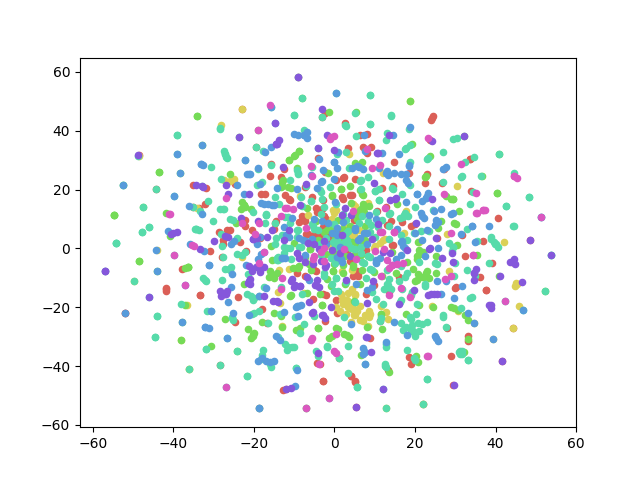}
\vspace{-22pt}
\caption*{(a) VAE}
\end{minipage}
\begin{minipage}[t]{0.32\textwidth}
\centering
\includegraphics[width=\textwidth]{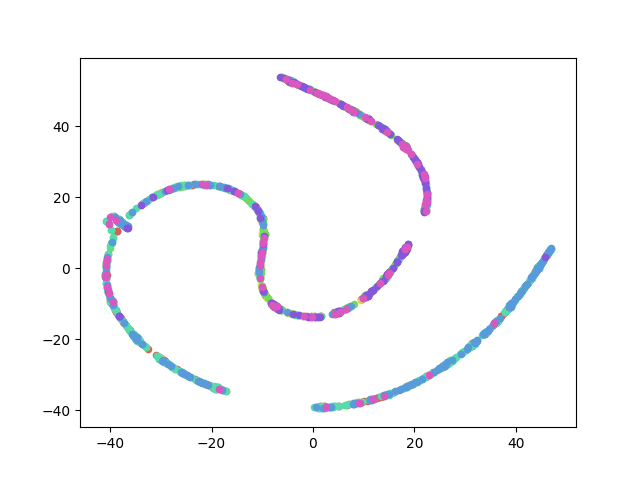}
\vspace{-22pt}
\caption*{(b) GCN}
\end{minipage}
\begin{minipage}[t]{0.32\textwidth}
\centering
\includegraphics[width=\textwidth]{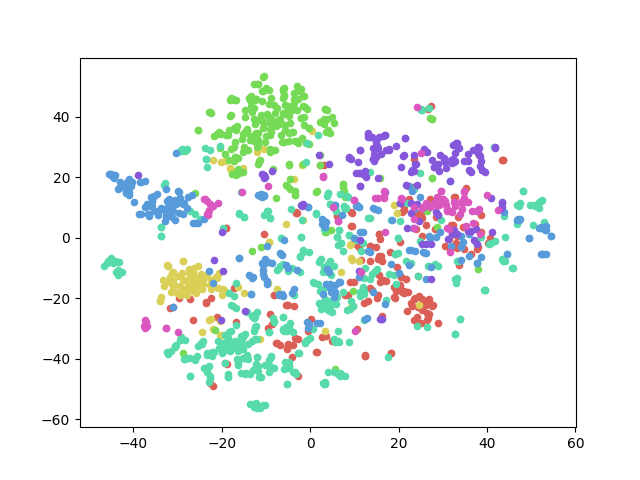}
\vspace{-22pt}
\caption*{(c) NANG}
\end{minipage}
\caption{The t-SNE visualization of test node embeddings on Cora. Each color represents one of seven classes. Note that all methods learn the node embeddings without class information.}
\label{figure:TSNE_result}
\end{figure*}

The involved methods in this paper, VAE, GCN and NANG encode node information into low-dimensional embedding and decode it as node attributes. Good representation ability means that a method can learn representative embedding where nearby nodes correspond to similar objects. Therefore, we conduct an experiment to visualize the learned node embeddings by t-SNE~\cite{maaten2008visualizing}. Specifically, the latent embeddings for all test nodes are sampled and we use t-SNE to make dimension reduction and visualize them in 2-D space on Cora dataset. Nodes in the same class are expected to be clustered together. Note that for all methods, they do not use label information in the training process. Therefore, the t-SNE visualization result is learned without class supervision for all methods. Figure~\ref{figure:TSNE_result} shows the corresponding result.

For VAE with Gaussian prior in Figure~\ref{figure:TSNE_result} (a), we can clearly see that the nodes of different classes are mixed together, which means it cannot distinguish the nodes belonging to different classes. For GCN in Figure~\ref{figure:TSNE_result} (b), it seems that the nodes are encoded into a narrow and stream like space, where different nodes are mixed and overlapped. Compared to VAE, the narrow and stream like space of GCN happens mainly because it has no prior assumption, which makes it lose distributed constraint. As for our NANG in Figure~\ref{figure:TSNE_result} (c), we can clearly see that different nodes are clustered well in accordance with their classes. Although Gaussian prior is both imposed on the coding space of VAE and NANG, our NANG can make information supplement between attribute modality and structure modality, yet capture more complex pattern for the latent space $Z$ while VAE fails. Therefore, NANG presents better t-SNE visualization result.

\begin{figure*}[t]
\centering
\begin{minipage}[t]{0.32\textwidth}
\centering
\includegraphics[width=\textwidth]{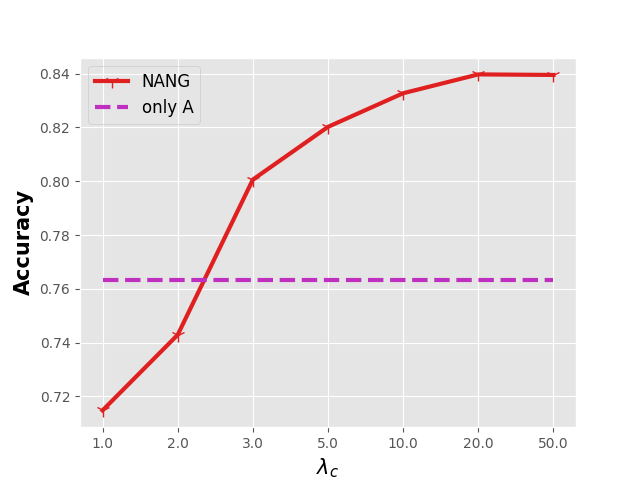}
\vspace{-15pt}
\caption*{(a) A+X - Cora}
\end{minipage}
\begin{minipage}[t]{0.32\textwidth}
\centering
\includegraphics[width=\textwidth]{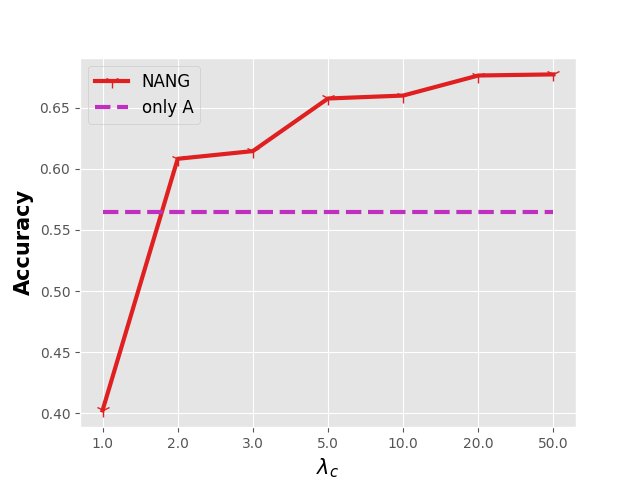}
\vspace{-15pt}
\caption*{(b) A+X - Citeseer}
\end{minipage}
\begin{minipage}[t]{0.32\textwidth}
\centering
\includegraphics[width=\textwidth]{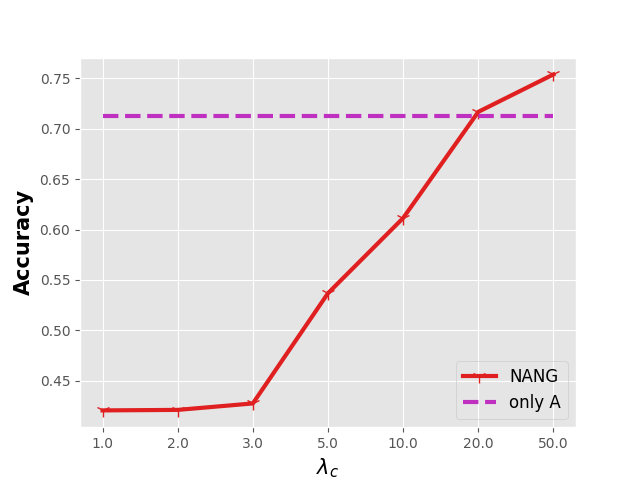}
\vspace{-15pt}
\caption*{(c) A+X - Pubmed}
\end{minipage} \\
\begin{minipage}[t]{0.32\textwidth}
\centering
\includegraphics[width=\textwidth]{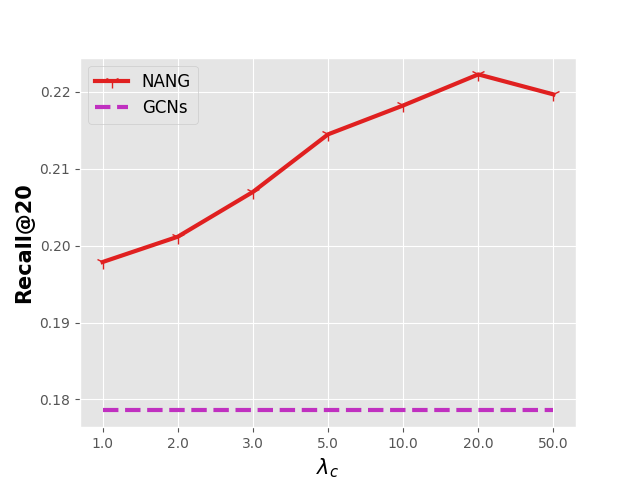}
\vspace{-15pt}
\caption*{(d) Recall - Cora}
\end{minipage}
\begin{minipage}[t]{0.32\textwidth}
\centering
\includegraphics[width=\textwidth]{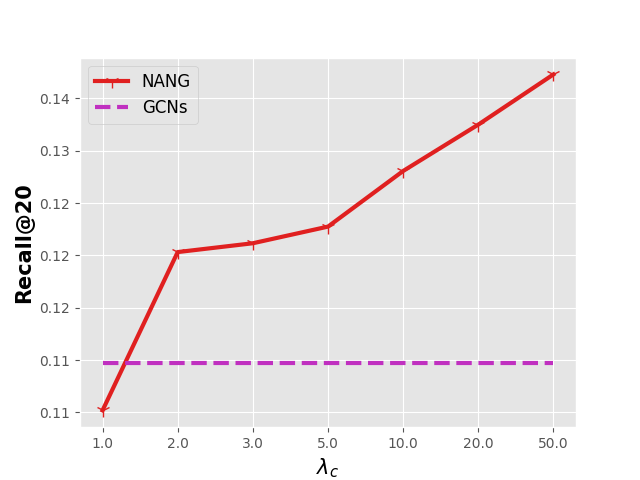}
\vspace{-15pt}
\caption*{(e) Recall - Citeseer}
\end{minipage}
\begin{minipage}[t]{0.32\textwidth}
\centering
\includegraphics[width=\textwidth]{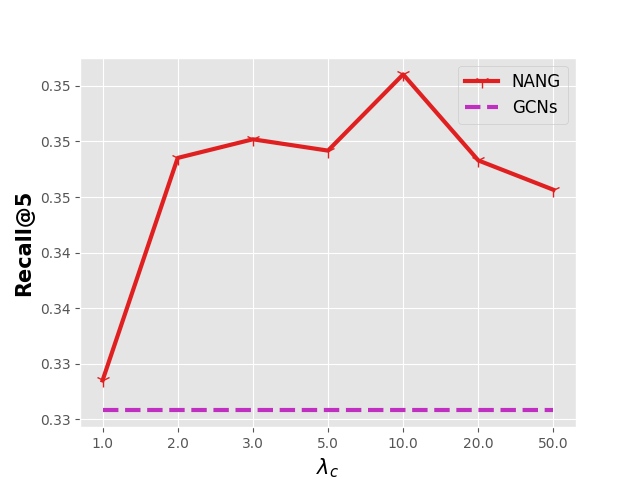}
\vspace{-15pt}
\caption*{(f) Recall - Steam}
\end{minipage}
\caption{NANG performance with different $\lambda_\mathrm{c}$ on both the node classification with "A+X" setting and profiling task. (a-c) means the result for node classification with "A+X" setting on Cora, Citeseer and Pubmed, respectively. The dotted line with "only A" represents that only the structure information is used, in which GCN as the classifier. (d-f) indicates the result for profiling on Cora, Citeseer and Steam, respectively. The dotted line with "GCN" means we use the GCN as the generation method because it is the most competitive baseline for profiling task in our paper.}
\label{figure:lambda_c}
\end{figure*}
\section[]{Hyperparameter $\lambda_\mathrm{c}$}
In our NANG, we introduce $\lambda_\mathrm{c}>1.0$ to emphasize the cross-reconstruction stream in our objective function. It is desirable to see how our method responds to this hyperparameter. Intuitively, we conduct an experiment about the node classification with "A+X" setting and profiling performance with different $\lambda_\mathrm{c}=[1.0,2.0,3.0,5.0,10.0,20.0,50.0]$. 

Figure~\ref{figure:lambda_c} (a-c) shows the result for node classification with "A+X" setting on Cora, Citeseer and Pubmed, respectively. And Figure~\ref{figure:lambda_c} (d-f) indicates the result for profiling on Cora, Citeseer and Steam, respectively. From these figures, we can clearly see that the hyperparameter $\lambda_\mathrm{c}$ is important for our method since we rely on the cross-reconstruction stream to generate node attributes. In Figure~\ref{figure:lambda_c} (a-c), it shows that we need a large $\lambda_\mathrm{c}$ to generate high-quality node attributes which could augment GCN classifier with only "A" is used. In Figure~\ref{figure:lambda_c} (d-f), NANG can mostly perform better than the most competitive baseline GCN in our paper. And for this result on Steam, too large $\lambda_\mathrm{c}$ could deteriorate the model performance because a large $\lambda_\mathrm{c}$ can weaken the importance of distribution matching. Therefore, $\lambda_\mathrm{c}$ should be chosen according to specific datasets.

\begin{figure*}[ht]
\centering
\begin{minipage}[t]{0.45\textwidth}
\centering
\includegraphics[width=\textwidth]{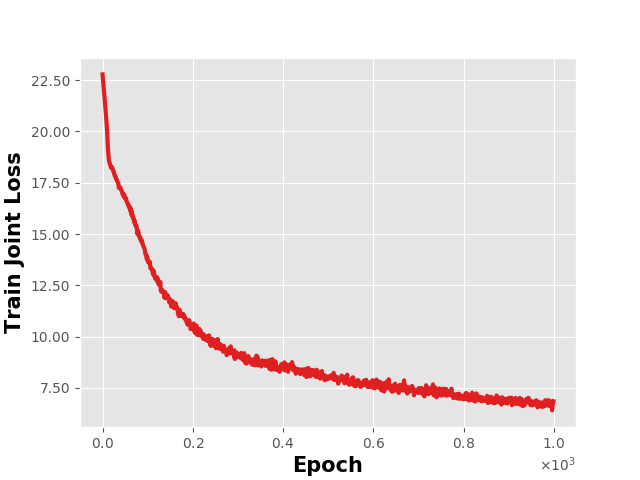}
\vspace{-15pt}
\caption*{(a) Train joint loss}
\end{minipage}
\begin{minipage}[t]{0.45\textwidth}
\centering
\includegraphics[width=\textwidth]{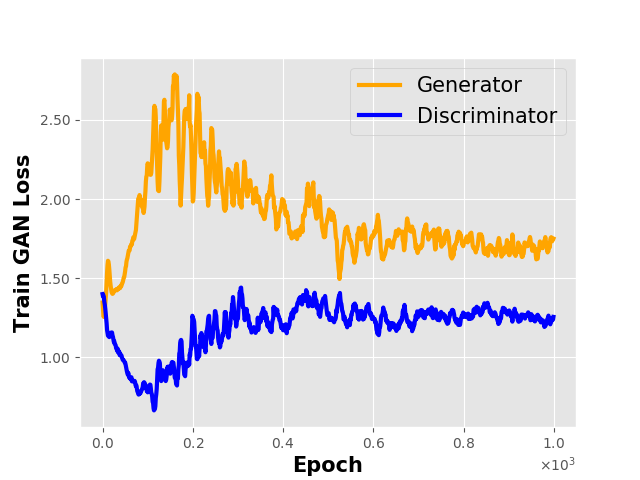}
\vspace{-15pt}
\caption*{(b) Train GANs loss}
\end{minipage} \\
\begin{minipage}[t]{0.45\textwidth}
\centering
\includegraphics[width=\textwidth]{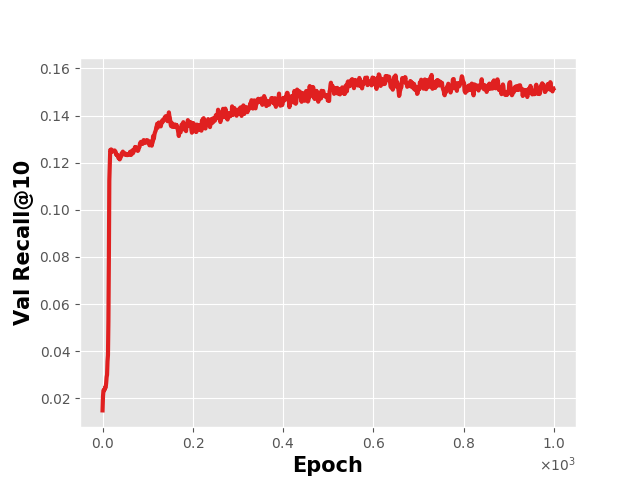}
\vspace{-15pt}
\caption*{(c) Validation metric}
\end{minipage}
\begin{minipage}[t]{0.45\textwidth}
\centering
\includegraphics[width=\textwidth]{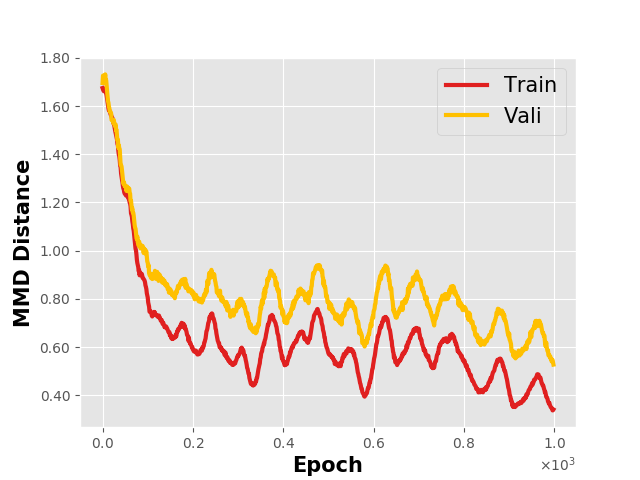}
\vspace{-15pt}
\caption*{(d) MMD distance}
\end{minipage}
\caption{Visualization of the training process for NANG on Cora. (a) The joint loss represents the sum of self-reconstruction stream and cross-reconstruction stream during the training process. (b) The GAN loss of the training process. (c) Validation Recall@10 along the training steps. (d) The train and validation MMD distance between learned $q(z)$ and Gaussian prior $p(Z)$.}
\label{figure:Train_process}
\end{figure*}
\section{Learning Process Visualization}
In order to understand the learning process of our method, we plot some learning curves including the train joint loss, train GAN loss, validation metric and MMD distance along the learning process. The result is indicated in Figure~\ref{figure:Train_process}.

This figure shows that both the train joint loss and train GAN loss converges, and the validation Recall@10 increases step by step and finally converges at around $800th$ epoch. The train and validation MMD distance is shown in Figure~\ref{figure:Train_process} (d), within the training process, two encoders $G_{X}$ and $G_{A}$ encode the input information and align them as the same latent factor whose distribution is an aggregated one $q(Z)$ from posteriors, the decreasing MMD distance indicates that the implicit distribution $q(Z)$ matches the whole distribution of $p(Z)$ step by step.

\end{document}